\pdfoutput=1

\documentclass[11pt]{article}

\usepackage{acl}

\usepackage{times}
\usepackage{latexsym}

\usepackage[T1]{fontenc}

\usepackage[utf8]{inputenc}

\usepackage{microtype}

\usepackage{inconsolata}

\usepackage{graphicx}
\usepackage{booktabs}
\usepackage{makecell}
\usepackage{adjustbox}
\usepackage{amsmath}
\usepackage{enumitem}
\usepackage{amssymb}
\usepackage{multirow}
\usepackage{wasysym}
\usepackage{float}
\usepackage{hyperref}

\title{Unleashing the Potentials of Likelihood Composition for Multi-modal Language Models}

\author{
 \textbf{Shitian Zhao\textsuperscript{1}}
 \textbf{Renrui Zhang\textsuperscript{1,2}}
 \textbf{Xu Luo\textsuperscript{1}}
 \textbf{Yan Wang\textsuperscript{3}}
 \\
 \textbf{Shanghang Zhang\textsuperscript{4}}
 \textbf{Peng Gao\textsuperscript{1}}
\\
\\
 \textsuperscript{1}Shanghai AI Laboratory
 \textsuperscript{2}CUHK
 \textsuperscript{3}East China Normal University
 \textsuperscript{4}Peking University
\\
}

\begin{document}
\maketitle
\begin{abstract}
 Model fusing has always been an important topic, especially in an era where large language models (LLM) and multi-modal language models (MLM) with different architectures, parameter sizes and training pipelines, are being created all the time. In this work, we propose a post-hoc framework, aiming at fusing heterogeneous models off-the-shell, which we call \textit{likelihood composition}, and the basic idea is to compose multiple models' likelihood distribution when doing a multi-choice visual-question-answering task. Here the core concept, \textit{likelihood}, is actually the log-probability of the candidate answer. In \textit{likelihood composition}, we introduce some basic operations: \textit{debias}, \textit{highlight}, \textit{majority-vote} and \textit{ensemble}. By combining (composing) these basic elements, we get the mixed composition methods: \textit{mix-composition}. Through conducting comprehensive experiments on 9 VQA datasets and 10 MLMs, we prove the effectiveness of \textit{mix-composition} compared with simple \textit{ensemble} or \textit{majority-vote} methods. In this framework, people can propose new basic composition methods and combine them to get the new mixed composition methods. We hope our proposed \textit{likelihood composition} can provide a new perspective of fusing heterogeneous models and inspire the exploration under this framework.\footnote{Code is released \href{https://github.com/zhaoshitian/Likelihood-Composition-Toolkit}{zhaoshitian/Likelihood-Composition-Toolkit}}
\end{abstract}

\section{Introduction}

Recently numerous multi-modal language models are emerging, \emph{e.g.},  LLaVA~\cite{liu2023llava,liu2023improvedllava,liu2024llavanext}, MiniGPT4~\cite{chen2023minigpt2}, BLIP-2~\cite{li2023blip2}, Qwen-VL~\cite{Qwen-VL}, InternVL~\cite{chen2023internvl}, and SPHINX~\cite{lin2023sphinx,gao2024sphinx_x}, each characterized by different architectures, parameter sizes, training datasets, and pipelines. Consequently, these models exhibit varying strengths across different tasks and domains. Some works~\cite{Gabriel23taskvector,wortsman2021wiseft,wortsman22modelsoup} have demonstrated that fusing multiple models can enhance performance and generalizability across diverse domains. Thus, several model fusion techniques have been devised to leverage the complementary capabilities of these models.

Many works focus on getting a new model by inheriting the knowledge from multiple parent models. Some of them interpolate several models' weights to get the new model's weight, \emph{e.g.}, WiSE-FT~\cite{wortsman2021wiseft} and model soup~\cite{wortsman22modelsoup}. However, in this process, all the parent models and the new derived model need to have the same architecture and parameter sizes, \emph{i.e.}, in most cases, the parent models are the fine-tuning versions of one pretrained model, leading to lack of diversity of these models. There are also some works focusing on distilling the knowledge from several different parent models~\cite{wan2024knowledge,wan2024fusechat}. However, the training computation cost makes it hard for researchers to combine parent models freely, \emph{i.e.}, the computation cost of the distillation training process limits researchers to do lots of trial-and-error experiments to get a good parent models recipe. 

Considering these issues, a promising line of work~\cite{zhao2023causalcog,li2024more,chuang2023dola,wang2022selfconsistent} fuse different models via manipulating or composing their likelihood distributions\footnote{Presicely, ``likelihood'' actually refers the log-probability of the generated answer, following~\cite{wang2022selfconsistent}, the detailed computing method is shown in Sec.\ref{sec:task_formulation}. Likelihood distribution contains likelihood of multiple candidate answers, here we assume by default that we are discussing a multiple-choice visual question task. }, with the advantage of being fully post-hoc, training-free and the same architecture of parent models is not necessary. The basic operation is to average all models' likelihood distribution of the candidate answers, called ``ensemble''.~\cite{dietterich2000ensemble} Recently some works also tried to combine the likelihood distributions from one model with different prompts as input~\cite{zhao2023causalcog,o2023contrastive,leng2023mitigating,zhang2023alleviating}. Behind these works, the core concept is to fuse different models via combining their likelihood distribution of the candidate answers, to boost the performance of downstream tasks, specifically multi-choice VQA tasks. Thus, we propose a framework named \textbf{likelihood composition}. Under this framework, we introduce some basic operations: \textit{debias}, \textit{highlight}, \textit{ensemble} and \textit{majority-vote}, some of which have been proposed in the previous research~\cite{niu2021counterfactual,zhao2023causalcog,wang2022selfconsistent}. And these basic operations are classified into two classes: ``self-composition'' and ``mutual-composition''. By composing these basic operations, we derive some new likelihood composition methods, \emph{e.g.}, ensemble-debias, ensemble-highlight, majority-debias and majority-highlight, which we call ``mix-composition''. 

To explore the likelihood composition's effectiveness on fusing models and boosting the performance on downstream tasks, we conduct the experiments on LLaVA series and other 4 advanced multi-modal language models and 9 VQA datasets. Our experiment results reveal some interesting findings:

\begin{figure}[t]
    \centering
    \includegraphics[width=\linewidth]{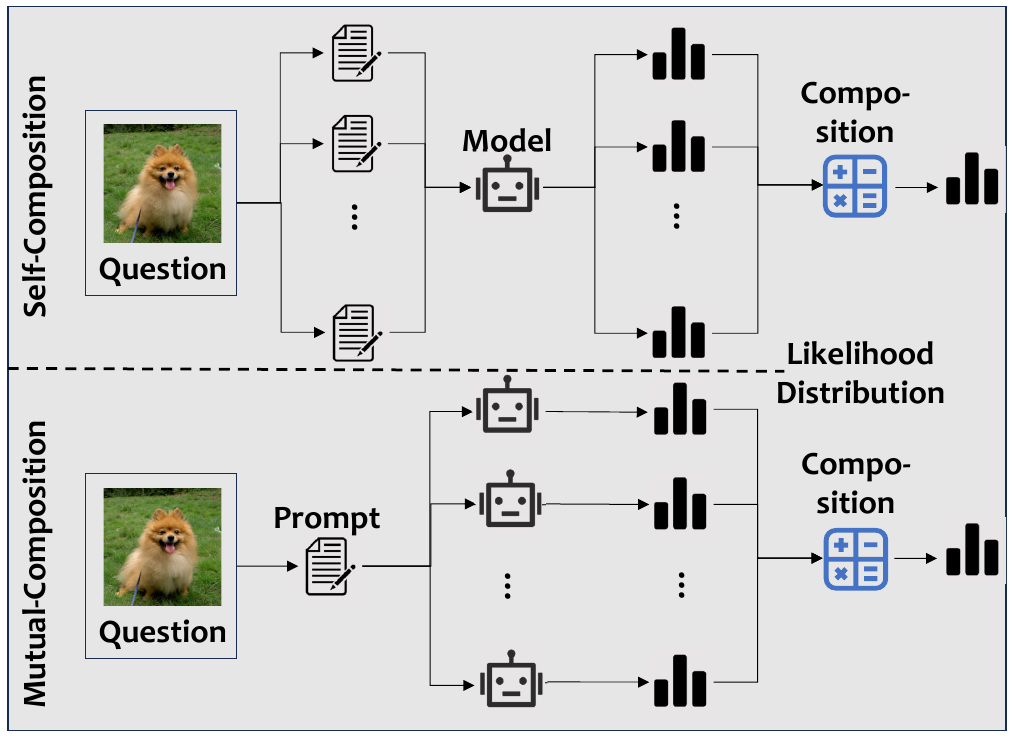}
    \caption{Two categories of likelihood composition: self-composition and mutual-composition.}
    \label{fig:framework}
\end{figure}

\begin{itemize}
\item [(1)] Self-composition can help model improve its performance on VQA tasks, especially for not well-developed models, \emph{e.g.}, \textit{debias} can bring a $+12.08\%$ improvement for LLaVA-7B on MMVP. 

\item [(2)] Mix-composition performs better compared to mutual-composition with respect to boosting the performance on VQA tasks, \emph{e.g.}, simply combining \textit{debias} and \textit{ensemble} can bring a $+7.93\%$ improvement on MMVP compared to vanilla ensemble method, a $+6.93$ improvement on MME compared to vanilla majority-vote. 

\item [(3)] When fusing models using likelihood composition, models' quality is more important than models' quantity, \emph{e.g.}, fusing LLaVA1.5-13B, LLaVA1.6-7B and LLaVA1.6-13B can make a better performance than fusing all models in LLaVA series.

\end{itemize}

\section{Related Works}

\noindent
\textbf{Multi-modal Language Models}
Based on the booming of large language models, lots of multi-modal language models have developed, \emph{e.g.}, LLaVA\cite{liu2023llava}, InternVL\cite{chen2023internvl}, Qwen-VL\cite{Qwen-VL}, Yi-VL, SPHINX\cite{gao2024sphinx_x} and CogAgent\cite{hong2023cogagent}. These models have similar architectures and training pipelines. The composition of the model architecture is to use a MLP layer or Q-Former\cite{li2023blip2} to connect a pretrained visual encoder, which could be the visual encoder in CLIP\cite{radford2021clip} or pretrained DINO\cite{caron2021dino}, to a pretrained large language model, \emph{e.g.}, LLaMA\cite{touvron2023llama2}, Mistral and InternLM\cite{team2023internlm}. The training pipeline mainly follows the two stage design: first align the vision and language modality by training on massive image-text pair data, \emph{e.g.}, CC3M and CC12M\cite{changpinyo2021cc12m}; then fine-tune the model using visual instruction data\cite{liu2023llava}. By doing so, these models show a good performance on multi-modal understanding and VQA tasks.

\vspace{1em}
\noindent
\textbf{Model Ensemble}
To boost the performance on downstream tasks, a usual method is to ensemble multiple models. In this literature, many lines of research are developed: weight interpolation\cite{wortsman22modelsoup,wortsman2021wiseft}, model collaboration\cite{besta2023graph,hong2023metagpt,shen2024hugginggpt,yao2023tree} and distillation-based methods\cite{wan2024fusechat,wan2024knowledge}.

\vspace{1em}
\noindent
\textbf{Decoding Methods for Language Models}
Given a pretrained large language model, the decoding method also matters a lot. Some works focus on the decoding techniques of LLM, \emph{e.g.}, contrastive decoding\cite{o2023contrastive}, contrasting a pair of weak and strong LLM's logits to improve the strong model's generation quality; proxy tuning\cite{liu2024proxytuning,liu2021dexperts}, doing arithmetic among three pretrained LLMs to boost the generation ability.

\section{Preliminary}

Before introducing the detailed methodology of likelihood composition officially, it is necessary to make clear some core concepts that may be mentioned in the methodology part. So in this section, we give a clear description of the task formulation, which is the basic setting of our study, and an accurate definition of ``likelihood'', the core concept used in our framework.

\noindent
\textbf{Task Formulation}
\label{sec:task_formulation}
Considering most multi-modal tasks, \emph{e.g.}, visual grounding and image retrieval can be formulated as a visual-question-answering task\cite{gao2024sphinx_x}. In this paper, we exclusively investigate multi-choice visual question answering (VQA) tasks. The task formulation is as follows: Given a dataset $\mathcal{D}=\{S_{i}\}$ comprising numerous VQA samples, $S_{i}=(\text{I}_{i}, \text{Q}_{i}, \textbf{C}_{i})$ consists of an image $\text{I}_{i}$ and a question $\text{Q}_{i}$. The candidate answers for this image-question pair are represented as $\mathbf{C}_{i}$, a list containing $n$ candidate answers: $\textbf{C}_{i}=[c_{0}^{i},..., c_{n}^{i}]$. And we need to input $(\text{I}_{i}, \text{Q}_{i}, \textbf{C}_{i})$ into MLM to predict the right answer's option letter.

\noindent
\textbf{Likelihood Calculation}
For a VQA sample ($\text{I},\text{Q},\mathbf{C}$), during the normal forward process, both $\text{I}$ and $\text{Q}$ are input into the model. $\mathbf{C}$ is a list of choices, denoted as $[c_{0}, c_{1}, ..., c_{n}]$, where $n$ is the number of choices. The likelihood of candidate $c_{i}$ is calculated as follows:

    \begin{equation}
    \begin{aligned}
        {y}_{i}(\text{X})=\exp^{\frac{1}{K_{i}}\sum_{k=1}^{K_{i}}\log P(t_{k}|\text{X},t_{1},t_{2},...,t_{k-1})},
    \end{aligned}
    \end{equation}
where $y_{i}$ represents the likelihood value of $c_{i}$ conditioned on $X$, the input to the model. $P(t_{k} | X, t_{1}, t_{2}, ..., t_{k-1})$ denotes the probability of generating the $k$th token $t_{k}$ conditioned on $X$ and the previously generated tokens $t_{1} \sim t_{k-1}$. $K_{i}$ represents the total number of tokens in $c_{i}$.

Once the likelihood of each choice is calculated, we denote the list containing all the choices' likelihood values as $\mathbf{Y}$, upon which composition is performed.\footnote{Here, after getting $\textbf{Y}$, we actually perform softmax on it to make likelihood distributions from different models at the same scale. For convenience, we use $\textbf{Y}$ to illustrate the likelihood composition framework and the experiments in the following sections.}\footnote{It should be noted that in our practice, the input $X$ also contains $\textbf{C}$, and each option's likelihood is actually the corresponding option letter's likelihood.}

\section{Likelihood Composition}

First, we organize the basic operations in the likelihood composition framework into two categories: \textit{self-composition} and \textit{mutual-composition}, as illustrated in Fig.\ref{fig:framework}. Then we mix these basic elements, resulting in the \textit{mix-composition}.

\subsection{Self-Composition}
\label{sec:self_composition}
The primary idea behind self-composition involves devising various prompt formats and preprocessing input samples using these prompts to generate different explicit inputs. Considering a VQA sample $S=(\text{I},\text{Q},\mathbf{C})$, we design $m$ prompting methods: $\text{Prompt}_{1}, ..., \text{Prompt}_{m}$. By applying these prompting methods to $S$, we obtain multiple results: $\text{X}_{1}, ..., \text{X}_{m}$. Subsequently, inputting these results into the model yields corresponding likelihood distributions on $\mathbf{C}$: $\textbf{Y}_{1}, ...,\textbf{Y}_{m}$, on which we conduct the composition method. In this section, we introduce two self-composition methods: \textit{Debias} and \textit{Highlight}.

\noindent
\textbf{Debias}

\begin{figure}[ht]
    \centering
    \includegraphics[width=0.9\linewidth]{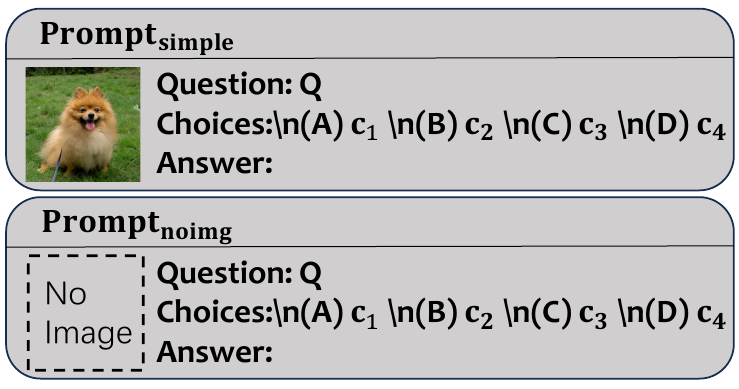}
    \caption{Prompt design of \textit{debias}.}
    \label{fig:prompt_debias}
    \vspace{-0.5em}
\end{figure}

\noindent
For multi-modal language models, language priors may be modeled during the training process on the multi-modal datasets, \emph{i.e.}, the model will give the hallucinated answer ignoring the actual content of the provided image.~\cite{agrawal2018don,li2023pope,niu2021counterfactual} For example, when providing a picture containing a black color banana to an MLM and ask ``What is the color of the banana?'', the model will say ``Yellow'', which is the language prior bias in the training set. To model the language prior bias existing in the MLM, we only input the question into the model, inducing the model to give the most common answer, which reflects the language bias. 

The prompt design is shown in Fig.\ref{fig:prompt_debias}.
Based on these prompt methods, we obtain corresponding likelihood values, $\textbf{Y}_{simple}$ and $\textbf{Y}_{noimg}$. We then subtract $\textbf{Y}_{noimg}$ from $\textbf{Y}_{simple}$ with a coefficient $\alpha$, as formalized below:
\begin{equation}
\small
\textbf{Y} = (1+\alpha)\textbf{Y}_{simple}-\alpha\textbf{Y}_{noimg}
\end{equation}
Finally, among the likelihood values of the $n$ choices in $\textbf{Y}$, we select the option corresponding to the highest likelihood as the predicted answer.


\noindent
\textbf{Highlight}

\begin{figure}[h]
    \centering
    \includegraphics[width=0.9\linewidth]{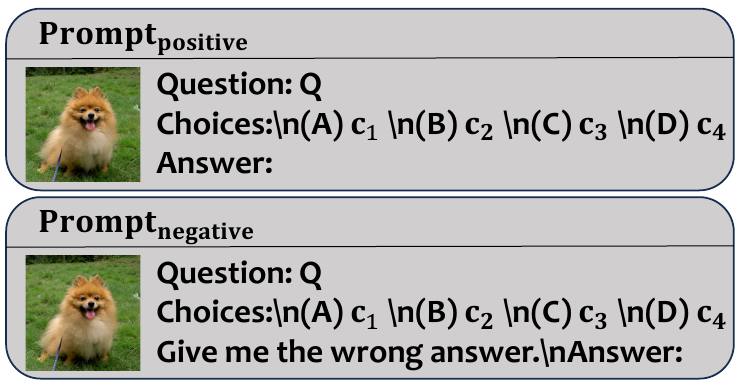}
    \caption{Prompt design of \textit{highlight}.}
    \label{fig:prompt_highlight}
    \vspace{-0.5em}
\end{figure}

\noindent
Based on the idea of highlighting by using the opposite side, we introduce \textit{Highlight}. First, we use the a positive instruction, \emph{e.g.}, ``Give me the right answer.'', to instruct the model to produce a high likelihood value of the right candidate. Then we use a negative instruction, \emph{e.g.}, ``Give me the wrong answer.'', to prone the model to the wrong candidates, producing a high likelihood value for the wrong answers. Finally, by contrasting these two likelihood distributions, we highlight the right answer.

We list the prompt design of highlight in Fig.\ref{fig:prompt_highlight}. It should be noted that in $\text{Prompt}_{positive}$ we do not use a positive instruction, \emph{e.g.}, ``Give me the right answer.'', since usually MLM will give the right answer with no special note.
Wrap the $(\text{I},\text{Q},\mathbf{C})$ with $\text{Prompt}_{negative}$, we get the explicit input to the model. Then, the corresponding likelihood, termed as $\textbf{Y}_{negative}$ is produced conditioned on the $\text{Prompt}_{negative}(\text{I},\text{Q},\mathbf{C})$. To highlight the right answer among all the candidates, we subtract $\textbf{Y}_{negative}$ from $\textbf{Y}_{positive}$:
\begin{equation}
\small
\textbf{Y}_{highlight} = (1+\alpha)\textbf{Y}_{positive}-\alpha\textbf{Y}_{negative},
\end{equation}
on which the selection of the predicted answer is based.

\subsection{Mutual-Composition}
\label{sec:mutual_composition}

Except for composing the likelihood distribution produced by one model, we can also compose the likelihood distribution output from multiple different models with varying architectures, sizes and training pipelines, termed as ``mutual-composition''. 

\noindent
\textbf{Ensemble}

\noindent
Based on the task formulation mentioned in Sec.~\ref{sec:task_formulation}, the most explicit composition method is to averaging all the provided likelihood distribution from different models, conditioned on the same input. Specifically, say there are $\text{N}$ models: $\{F_{i}|i=1,...,\text{N}\}$, and conditioned on the given sample, $(\text{I},\text{Q},\mathbf{C})$, we can get $\text{N}$ likelihood distribution of the candidate answers from the $\text{N}$ corresponding models, termed as $\{\textbf{Y}_{i}|i=1,...,\text{N}\}$. To ensemble them, we do the simplest averaging:
\begin{equation}
\small
    \textbf{Y}_{ensemble} = \frac{1}{\text{N}}\sum_{i=1}^{\text{N}}\textbf{Y}_{i},
\end{equation}
from which the predicted option is selected.

\noindent
\textbf{Majority-vote}

\noindent
There have been some works focusing on ensemble models' output, \emph{e.g.}, CoT-SC\cite{wang2022selfconsistent}. The basic operation used in these works is to do the majority-voting among the outputs despite being produced by one model or multiple heterogeneous models. 
And majority-vote actually can also be expressed using the likelihood composition language. Compared to ensemble mentioned above, majority-vote actually just adds a mask during the composition process:

\begin{equation}
\small
\left\{
\begin{aligned}
& \textbf{Y}_{majority-vote}=\frac{1}{\text{N}}\sum_{i=1}^{\text{N}}\textbf{1}*[MASK]_{i}, &\text{Unweighted} \\
& \textbf{Y}_{majority-vote}=\frac{1}{\text{N}}\sum_{i=1}^{\text{N}}\textbf{Y}_{i}*[MASK]_{i}, &\text{Weighted},
\end{aligned}
\right.
    \end{equation}
where \textbf{1} is an all ones vector having the same length with $\textbf{Y}_{i}$. $[MASK]_{i}$ is a 0-1 vector, where the element with the same index as that of the max element in $\textbf{Y}_{i}$ is 1, and the other elements are 0.

\subsection{Mix the Self-Composition and Mutual-Composition}
The basic idea of likelihood composition is to compose different likelihood distributions from either one model or multiple heterogeneous models. From this perspective, we can mix the two introduced composition methods mentioned in Sec.~\ref{sec:self_composition} and Sec.~\ref{sec:mutual_composition}, and we call the mixed method: \textbf{mix-composition}.  

In detail, say there are \text{N} models, $\{\text{F}_{i}| i=1,...,\text{N}\}$ we first apply the self-composition, either \textit{debias} or \textit{highlight}, to the likelihood distribution output by one model. Then, we use the \textit{ensemble} method to fuse the \text{N} manipulated likelihood distribution of different models. The arithmetic formula is shown below:
\begin{equation}
\small
    \textbf{Y}_{mix}=\frac{1}{\text{N}}\sum_{i=1}^{N}((1+\alpha)\textbf{Y}_{simple,i} - \alpha\textbf{Y}_{*,i}),
\end{equation}
where * is either \textit{noimg} or \textit{negative}, the prompt method used in self-composition.

Similarly, when combining self-composition and majority-vote, the composition would be:
\begin{equation}
\small
\textbf{Y}_{mix}=\frac{1}{\text{N}}\sum_{i=1}^{\text{N}}((1+\alpha)\textbf{Y}_{simple,i} - \alpha\textbf{Y}_{*,i})*[MASK]_{i}, 
\end{equation}
where * is either \textit{noimg} or \textit{negative}.

\section{Experiments}

To evaluate the efficiency of likelihood composition methods on improving MLM's performance on VQA tasks, we conduct experiments on 10 advanced multi-modal language models and 9 VQA benchmarks.

\subsection{Used Multi-modal Language Model}
In our early experiments, we find the likelihood composition's efficiency varied in models with different levels of capabilities and different training schemas. To show the likelihood composition's general effectiveness, we select the LLaVA series, from LLaVA-7B to LLaVA1.6-13B, with similar architectures, training schemas and stepwise capability increase. Also, we select other 4 well-developed MLMs: Yi-VL, Qwen-VL, InternVL, and Internlm-Xcomposer, to see how likelihood composition performs on heterogeneous models.
\begin{itemize}
\small
    \item [-] \textbf{LLaVA series}\cite{liu2023llava,liu2023improvedllava} are multi-modal language models developed on pretrained large language models, \emph{e.g.}, LLaMA\cite{touvron2023llama2} and Vicuna\cite{vicuna2023}, and pretrained visual encoder, \emph{e.g.}, vision encoder in pretrained CLIP\cite{radford2021clip}. A simple linear layer or MLP layer is used for aligning the vision and language modalities. Models in this family are trained on vision-text pairwise data and visual instruction data. We select LLaVA-7B, LLaVA-13B, LLaVA1.5-7B, LLaVA1.5-13B, LLaVA1.6-7B and LLaVA1.6-13B with increased multi-modal ability in our experiments.
    \item [-] \textbf{Other Heterogeneous MLMs} contains Yi-VL, Qwen-VL\cite{Qwen-VL}, InternVL\cite{chen2023internvl}, and Internlm-Xcomposer\cite{zhang2023internlmxcomposer}, which are all trained on vision-language pairwise data and multi-modal instruction data. These models have good visual understanding and reasoning abilities.
\end{itemize}

\subsection{Used Datasets}
We include different types of VQA datasets in our experiments, to prove the likelihood composition's generalizability with respect to data and tasks.

\begin{itemize}
\small
    \item [-] \textbf{Comprehensive VQA benchmarks} include MME\cite{fu2023mme} and MMBench\cite{liu2023mmbench}, two comprehensive VQA benchmarks containing various tasks, from object existence to code reasoning. And the question format is multi-choice VQA, which is fitted for our task formulation, introduced in Sec.~\ref{sec:task_formulation}.
    \item [-] \textbf{Diagnose VQA Benchmarks} contains VSR, POPE\cite{li2023pope} and MMVP\cite{tong2024mmvp}. POPE is a Yes-No format VQA benchmark aimed at diagnosing MLM's object hallucination. MMVP is a dataset exploring the shortcomings of MLMs.
    \item [-] \textbf{Reformed Academic VQA Datasets} include the split and reformed versions of VQAv2\cite{goyal2017vqav2}, OKVQA\cite{marino2019okvqa}, GQA\cite{hudson2019gqa} and Vizwiz\cite{gurari2018vizwiz} from ReForm-Eval\cite{li2023reform}, and we use a superscript ``*'' to represent the reformed versions. 
\end{itemize}

\begin{table*}[th]
\vspace{-1em}
    \centering
    \adjustbox{max width=0.9\textwidth}{
    \begin{tabular}{l|c|cc|ccccccccc}
    \toprule
    &$\alpha$&\makecell{De\\bias}&\makecell{High\\light}&MME&MMVP&MMBench&VSR&POPE&VQAv2*&Vizwiz*&GQA*&OKVQA* \\
    \midrule
         \multirow{5}{*}{7B}
         &&&&991.26&0.67&40.62&54.91&61.83&38.99&36.66&36.83&31.94 \\
         \cline{2-13}
         &\multirow{2}{*}{1.0}
         &\checkmark&&\textbf{1069.56}&\textbf{12.75}&44.67&57.20&\textbf{76.76}&\textbf{42.07}&\textbf{38.52}&\textbf{38.19}&\textbf{32.54}  \\
         &&&\checkmark&973.94&4.70&\textbf{46.70}&51.47&70.82&38.53&32.02&35.24&31.15  \\
         \cline{2-13}
         
         &\multirow{2}{*}{0.1}
         &\checkmark&&988.63&2.01&41.81&55.65&61.76&39.69&36.89&37.07&32.34 \\
         &&&\checkmark&987.84&4.03&42.86&\textbf{59.33}&63.35&39.97&36.43&36.20&31.94 \\
         \midrule
         \multirow{5}{*}{13B}
        &&&&1106.00&4.70&41.58&61.62&55.72&37.87&31.79&38.58&28.77 \\
        \cline{2-13}
        &\multirow{2}{*}{1.0}
         &\checkmark&&1124.50&8.72&38.50&\textbf{62.60}&\textbf{59.46}&38.20&\textbf{32.95}&37.07&29.17 \\
         &&&\checkmark&\textbf{1197.31}&\textbf{13.42}&40.16&55.32&54.64&34.10&25.06&33.09&25.00  \\
         \cline{2-13}
         &\multirow{2}{*}{0.1}
         &\checkmark&&1090.73&8.72&\textbf{45.97}&61.54&57.00&\textbf{39.46}&34.11&\textbf{38.98}&\textbf{30.56} \\
         &&&\checkmark&1114.30&8.72&41.88&60.31&58.83&36.99&31.32&36.04&27.78 \\
         \midrule
         \multirow{5}{*}{v1.5-7B}
        &&&&1741.14&24.16&71.17&58.92&85.78&73.09&64.04&65.08&73.81 \\
        \cline{2-13}
        &\multirow{2}{*}{1.0}
         &\checkmark&&1723.09&\textbf{25.50}&70.30&\textbf{65.88}&70.19&\textbf{73.60}&63.34&64.52&73.41  \\
         &&&\checkmark&\textbf{1804.74}&21.48&68.52&54.34&73.82&71.83&61.95&64.52&68.85 \\
         \cline{2-13}
&\multirow{2}{*}{0.1}
         &\checkmark&&1747.96&24.83&\textbf{71.42}&60.80&\textbf{85.95}&73.13&\textbf{64.27}&\textbf{65.31}&\textbf{73.81} \\
         &&&\checkmark&1756.19&22.82&71.05&58.02&73.43&73.13&64.04&65.08&72.22 \\
         \midrule
         \multirow{5}{*}{v1.5-13B}
        &&&&1782.34&26.17&73.09&68.17&84.70&75.70&75.17&67.70&76.19 \\
        \cline{2-13}
        &\multirow{2}{*}{1.0}
         &\checkmark&&\textbf{1833.70}&26.17&\textbf{73.25}&\textbf{73.90}&79.83&75.42&\textbf{75.64}&67.70&75.40  \\
         &&&\checkmark&1819.68&26.17&68.61&71.11&59.11&75.79&75.17&66.27&72.22  \\
         \cline{2-13}
         &\multirow{2}{*}{0.1}
         &\checkmark&&1789.38&\textbf{26.85}&73.22&69.89&\textbf{86.04}&75.75&75.41&67.70&75.79 \\
         &&&\checkmark&1779.75&25.50&72.97&70.29&60.56&\textbf{75.84}&75.14&\textbf{68.10}&\textbf{76.19} \\
         \midrule
         \multirow{5}{*}{v1.6-7B}
        &&&&1691.81&13.42&\textbf{71.30}&66.12&67.33&67.07&54.52&59.35&\textbf{72.62} \\
        \cline{2-13}
        &\multirow{2}{*}{1.0}
         &\checkmark&&\textbf{1765.97}&14.77&70.46&64.81&71.11&66.84&54.76&58.31&71.43  \\
         &&&\checkmark&1679.07&\textbf{17.45}&51.61&60.23&60.61&67.07&52.90&\textbf{59.51}&68.85  \\
         \cline{2-13}
         &\multirow{2}{*}{0.1}
         &\checkmark&&1711.07&13.42&71.14&\textbf{66.45}&\textbf{68.50}&67.16&54.99&59.11&72.42 \\
         &&&\checkmark&1703.37&14.09&70.41&65.96&72.42&\textbf{69.26}&\textbf{67.02}&59.19&72.42\\
         \midrule
         \multirow{5}{*}{v1.6-13B}
        &&&&1807.45&26.85&74.05&66.94&84.74&76.21&\textbf{81.67}&66.75&75.99 \\
        \cline{2-13}
        &\multirow{2}{*}{1.0}
         &\checkmark&&1790.79&\textbf{28.86}&73.95&\textbf{69.89}&81.36&75.70&78.42&67.46&75.60  \\
         &&&\checkmark&1726.69&27.52&70.37&68.33&56.83&75.89&77.73&\textbf{68.18}&74.60  \\
         \cline{2-13}
         &\multirow{2}{*}{0.1}
         &\checkmark&&1800.68&28.19&\textbf{74.11}&67.92&\textbf{84.78}&\textbf{76.35}&80.97&67.06&\textbf{76.59} \\
         &&&\checkmark&\textbf{1814.42}&28.86&74.05&68.90&59.47&76.21&81.21&66.91&76.19 \\
    \bottomrule
    \end{tabular}}
    \caption{Self-composition methods bring a consistent improvement to all the LLaVA series models. (1) For \textit{debias}, when $\alpha$ is set to 1.0, it improves LLaVA-7B's performance on all 9 datasets and LLaVA-13B's performance on 7 datasets. When $\alpha$ is set to 0.1, \textit{debias} improves LLaVA1.5-7B and LLaVA1.5-13B's performance on 9 and 8 datasets respectively; improves LLaVA1.6-7B and LLaVA1.6-13B's performance on 5 and 7 datasets respectively. (2) For \textit{highlight}, when setting $\alpha$ to 0.1, it improves LLaVA1.6-13B, LLaVA1.6-7B and LLaVA-7B's performance on 5 datasets.}
    \label{tab:llava_series_self}
    \vspace{-1em}
\end{table*}

\vspace{-1em}
\subsection{Main Results}

We reported the results of applying likelihood composition on LLaVA series MLMs and 4 advanced MLMs with varying hyperparameters in Table.\ref{tab:llava_series_self},\ref{tab:llava_series_mutual} and Table.\ref{tab:other_mlm} respectively. 
In Table.\ref{tab:llava_series_self}, baseline is the MLM's intrinsic performance, while in Table.\ref{tab:llava_series_mutual} and Table.\ref{tab:other_mlm}, baselines are \textit{ensemble} and \textit{majority-vote}, which we refer as \textit{mutual-composition} in our framework.

\noindent
\textbf{Results on LLaVA Series}
As shown in Table.\ref{tab:llava_series_self}, applied with self-composition methods mentioned in Sec.~\ref{sec:self_composition}, LLaVA series' performance on the 9 datasets consistently improved, \emph{e.g.}, $+12.08\%$ for LLaVA-7B on MMVP, $+4.39\%$ for LLaVA-13B on MMBench, $+6.96\%$ for LLaVA1.5-7B on VSR, etc. Overall, for the early models in LLaVA family, \emph{i.e.}, LLaVA-7B and LLaVA-13B, which is not well developed relatively, self-composition methods improve their performance significantly. Also, aggressive self-composition, \emph{i.e.}, with $\alpha=1.0$ works better in most cases than that with $\alpha=0.1$, for LLaVA-7B and LLaVA-13B.

\begin{table*}[th]
    \centering
    \adjustbox{max width=0.9\textwidth}{
    \begin{tabular}{l|c|cc|ccccccccc}
    \toprule
    &$\alpha$&\makecell{De\\bias}&\makecell{High\\light}&MME&MMVP&MMBench&VSR&POPE&VQAv2*&Vizwiz*&GQA*&OKVQA* \\
    \midrule
         
    \multirow{12}{*}{Majority-vote}
    &\multicolumn{3}{c|}{unweighted}&1751.98&14.09&73.34&67.35&84.77&73.69&68.68&63.64&76.19 \\
    &\multicolumn{3}{c|}{weighted}&1826.62&23.49&74.48&69.64&\textbf{87.49}&77.80&74.48&69.21&79.17 \\
    \cline{2-13}
    &\multirow{3}{*}{1.0}
    &\checkmark&&1820.95&\textbf{36.91}&74.71&70.87&82.49&77.66&72.62&69.85&78.37 \\
    &&&\checkmark&1773.69&24.16&72.72&59.00&61.83&77.66&\textbf{75.17}&68.66&78.77 \\
    &&\checkmark&\checkmark&1797.63&30.20&74.37&67.76&70.05&77.99&74.48&69.29&79.17 \\
    \cline{2-13}
    &\multirow{3}{*}{0.5}
    &\checkmark&&1833.01&\underline{34.23}&74.87&\textbf{71.77}&85.35&78.08&72.85&70.01&78.17 \\
    &&&\checkmark&1833.32&24.16&73.22&63.18&62.26&77.94&\underline{74.94}&68.89&79.56 \\
    &&\checkmark&\checkmark&1797.60&28.19&74.39&69.56&76.53&77.85&74.25&69.93&78.97 \\
    \cline{2-13}
    &\multirow{3}{*}{0.1}
    &\checkmark&&\underline{1836.29}&26.17&74.53&70.30&\underline{87.30}&78.03&74.25&69.93&78.57 \\
    &&&\checkmark&1840.52&23.49&74.34&68.99&62.90&78.13&73.55&69.05&\textbf{80.56} \\
    &&\checkmark&\checkmark&1837.14&23.49&74.41&69.89&71.67&78.26&74.48&69.77&\underline{79.96} \\
    \midrule
    \multirow{11}{*}{Ensemble}
    &&&&1837.33&26.17&74.66&70.13&82.72&77.75&72.85&70.80&78.57\\
    \cline{2-13}
    &\multirow{3}{*}{1.0}
    &\checkmark&&1813.03&31.54&74.78&70.87&79.79&77.80&72.85&70.49&78.37 \\
    &&&\checkmark&1822.79&24.16&71.90&56.87&61.29&78.17&74.01&69.61&78.57 \\
    &&\checkmark&\checkmark&1832.44&27.52&73.79&66.12&61.82&78.26&73.78&69.69&78.17 \\
    \cline{2-13}
    &\multirow{3}{*}{0.5}
    &\checkmark&&1817.50&33.56&\textbf{74.94}&70.87&80.90&\textbf{78.36}&73.78&70.88&78.37 \\
    &&&\checkmark&1816.52&24.16&73.25&62.77&61.63&77.66&73.32&69.69&78.57 \\
    &&\checkmark&\checkmark&1823.61&27.52&74.37&67.84&62.36&78.22&74.48&70.41&78.97 \\
    \cline{2-13}
    &\multirow{3}{*}{0.1}
    &\checkmark&&\textbf{1839.75}&28.86&74.82&70.79&82.45&77.80&73.55&\underline{71.12}&78.37 \\
    &&&\checkmark&1836.15&26.17&74.48&69.72&66.51&77.52&73.09&70.56&78.57 \\
    &&\checkmark&\checkmark&1834.62&26.17&\underline{74.87}&70.54&80.25&77.80&73.32&70.64&78.37 \\
    \cline{2-13}
    &\multicolumn{3}{c|}{\makecell{$\alpha_{d}=0.5$ \\$\alpha_{h}=0.1$}}&1822.77&28.19&74.78&\underline{70.95}&83.35&\underline{78.26}&74.25&\textbf{71.20}&78.57 \\
    \bottomrule
    \end{tabular}}
    \caption{Combining self-composition and mutual-composition can significantly further improve the performance over that of mutual-composition. In most cases, combining \textit{debias} with mutual-composition can bring a further improvement. (1) When setting $\alpha$ to 0.5, combining \textit{debias} with \textit{majority-vote} can bring improvement on 6 datasets. (2) When setting $\alpha$ to 0.1, combining \textit{debias} with \textit{ensemble} can bring improvement on 7 datasets. (3) In the last line, we combine both \textit{debias} and \textit{highlight} and set the $\alpha_{d}$ to 0.5, $\alpha_{h}$ to 0.1.}
    \label{tab:llava_series_mutual}
    \vspace{-1em}
\end{table*}

For those well-developed models, \emph{i.e.}, LLaVA1.5-7B, LLaVA1.5-13B, LLaVA1.6-7B and LLaVA1.6-13B, the improvement self-composition brings is not as significant as before. In more detail, for the best model, LLaVA1.6-13B, the improvement on MMVP is $+2.01\%$, relatively small than that of LLaVA-7B: $+12.08\%$. In some cases, self-composition will cause the performance drop, \emph{e.g.}, the performance of LLaVA1.6-7B on MMBench dropped from $71.30\%$ to $71.14\%$. Concerning the $\alpha$, the value of 0.1 works better.

In Table.\ref{tab:llava_series_mutual}, we reported the results applying mutual-composition and mix-composition. For the mutual-composition, \emph{i.e.}, vanilla ensemble and weighted majority-vote (likelihood as the weight), the performance is significantly higher than that of unweighted majority-vote, which is a mainstream model ensemble and collaboration method. For example, on MMVP, the performance of the vanilla ensemble method is higher than that of unweighted majority-vote by $+12.08\%$ and this number for weighted majority-vote is $+9.4\%$. For mix-composition methods, which means mix the two self-compositon methods: \textit{debias} or \textit{highlight} into the mutual-composition pipeline, we can see that after the mixing, the performance on most VQA datasets will be improved further, \emph{e.g.}, mixing debias and weighted majority-vote brings a $+13.42\%$ improvement on MMVP and $+2.13\%$ improvement on VSR.

In general, by conducting experiments on model families with increased abilities, we find:
\begin{itemize}
    \item [-] Self-composition with a high $\alpha$ value works better for not well-developed models. While self-composition with a low $\alpha$ value is suitable for advanced models. More analysis could be found in Sec.\ref{sec:alpha's value}.
    \item [-] Mix-composition works better than mutual-composition when fusing different models.
\end{itemize}

\begin{table*}[]
\vspace{-1em}
    \centering
    \adjustbox{max width=0.9\textwidth}{
    \begin{tabular}{l|c|c|cccccccc}
    \toprule
&$\alpha$&\makecell{De\\bias}&MME&MMVP&VSR&POPE&VQAv2*&Vizwiz*&GQA*&OKVQA*\\
    \midrule
LLaVA&&&1807.45&26.85&66.94&84.74&76.21&81.67&66.75&75.99 \\  
        \midrule
 Yi-VL&&&1977.21&35.57&59.90&81.50&76.96&72.39&72.39&78.57  \\
        \midrule
Qwen-VL&&&1769.19&26.85&69.80&86.96&76.12&63.80&70.41&74.80  \\
        \midrule
InternVL&&&1714.89&17.45&69.39&85.97&60.12&46.87&55.93&37.10  \\
        \midrule
\makecell[l]{Internlm\\-Xcomposer}&&&1896.67&28.86&77.58&87.47&63.95&56.84&57.84&51.79  \\
        \midrule
\multirow{5}{*}{Ensemble}
    &&&2026.23&\textbf{38.93}&77.74&87.13&\textbf{81.58}&78.65&74.14&79.76 \\
    \cline{2-11}
    &1.0&\checkmark&2021.66&32.89&76.35&86.02&81.06&74.71&74.46&\textbf{80.56} \\
    \cline{2-11}
    &0.5&\checkmark&2009.15&36.24&76.51&\textbf{87.87}&\textbf{81.58}&76.8&74.46&80.16 \\
    \cline{2-11}
    &0.1&\checkmark&\textbf{2035.31}&35.57&77.66&87.63&81.53&78.65&\textbf{74.54}&80.16 \\
    \cline{2-11}
    &0.05&\checkmark&2030.53&37.58&\textbf{77.58}&87.34&81.53&\textbf{79.12}&74.38&79.76  \\
        \bottomrule
    \end{tabular}}
    \caption{Results of applying \textit{mutual-composition} and \textit{mix-composition} on LLaVA, Yi-VL, Qwen-VL, InternVL and Internlm-Xcomposer. As we can see, in most cases, mixing \textit{debias} with \textit{ensemble} can bring the further improvement.}
    \label{tab:other_mlm}
    \vspace{-1.2em}
\end{table*}

\vspace{-0.7em}
\noindent
\textbf{Results on 5 Advanced MLMs}
We apply \textit{mutual-composition} and \textit{mix-composition} on 5 advanced MLMs: LLaVA, Yi-VL, Qwen-VL, InternVL and Internlm-Xcomposer. As shown in Table.\ref{tab:other_mlm}, \textit{mutual-composition} brings significant improvement on most datasets, \emph{e.g.}, $+4.62\%$ improvement on VQAv2* and $+3.36\%$ improvement on MMVP. And \textit{mix-composition} brings further improvement, \emph{e.g.}, $+9.08$ on MME.

\vspace{-0.5em}
\section{Additional Analysis}

\begin{figure}[th]
    \centering
    \includegraphics[width=0.95\linewidth]{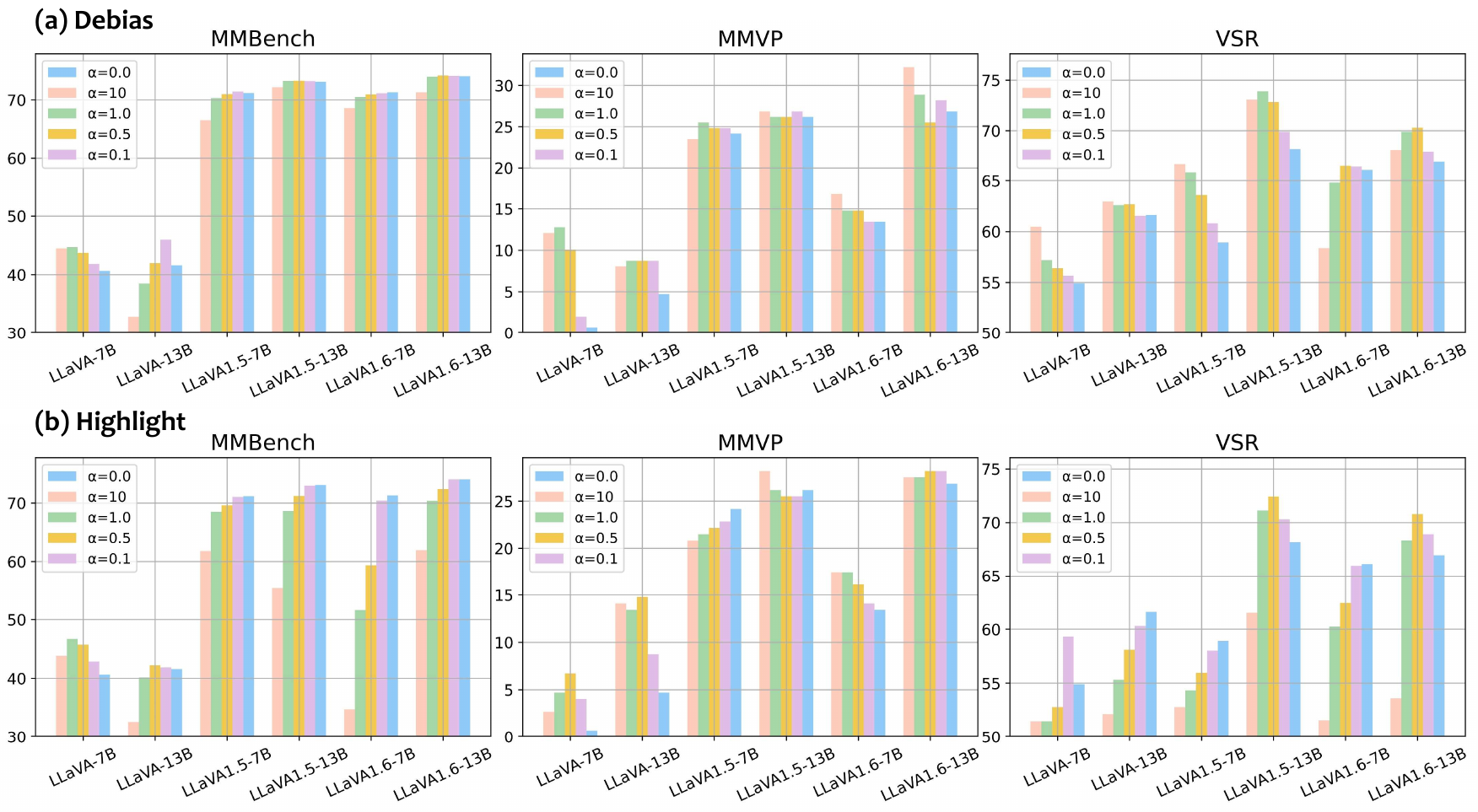}
    \caption{(a) Applying \textit{debias} on LLaVA series models with $\alpha$ ranging from 1.0 to 0.1. (b) Applying \textit{highlight} on LLaVA series models with $\alpha$ ranging from 1.0 to 0.1.}
    \label{fig:alpha}
    \vspace{-1em}
\end{figure}

\textbf{Exploring $\alpha$'s Consequence on Self-Composition.}
\label{sec:alpha's value}
In the previous experiments, $\alpha$'s value shows different consequences on models with different levels of ability. \textbf{Generally, a high $\alpha$'s value works better for not well-developed models and a low $\alpha$'s value is suitable for advanced models.} So we conduct a more detailed analysis, ranging the $\alpha$'s value from 0.1 to 1.0, and visualize the results in Fig.\ref{fig:alpha}.

When setting $\alpha=0.0$, we do not use the self-composition method, which is the baseline. In subfigure (a), as we can see, \textit{debias} works for all models on VSR. The highest results appear when $\alpha$ is set low for LLaVA-7B, LLaVA-13B, LLaVA1.5-7B and LLaVA1.5-13B, while for LLaVA1.6-7B and LLaVA1.6-13B, the best performance appears when $\alpha$ is set relatively low, which is 0.5. On MMBench and MMVP, \textit{debias} works well for LLaVA-7B and LLaVA-13B. But for the other more advanced models, a high $\alpha$ may bring damage to the performance and $\alpha$ with low value may bring improvement.\footnote{On MMVP, $\alpha$ with 1.0 works well for LLaVA1.6-7B and LLaVA1.6-13B, which are two relatively advanced model. But it should be noted that all models do not perform well on MMVP, \emph{i.e.}, on MMVP, all models are actually not ``advanced''.} In subfigure (b), \textit{highlight} does not work so well as \textit{debias}. But overall, \textit{highlight} works better for LLaVA-7B and LLaVA-13B, which are two relatively weak models. 



\begin{figure}[th]
    \centering
    \includegraphics[width=\linewidth]{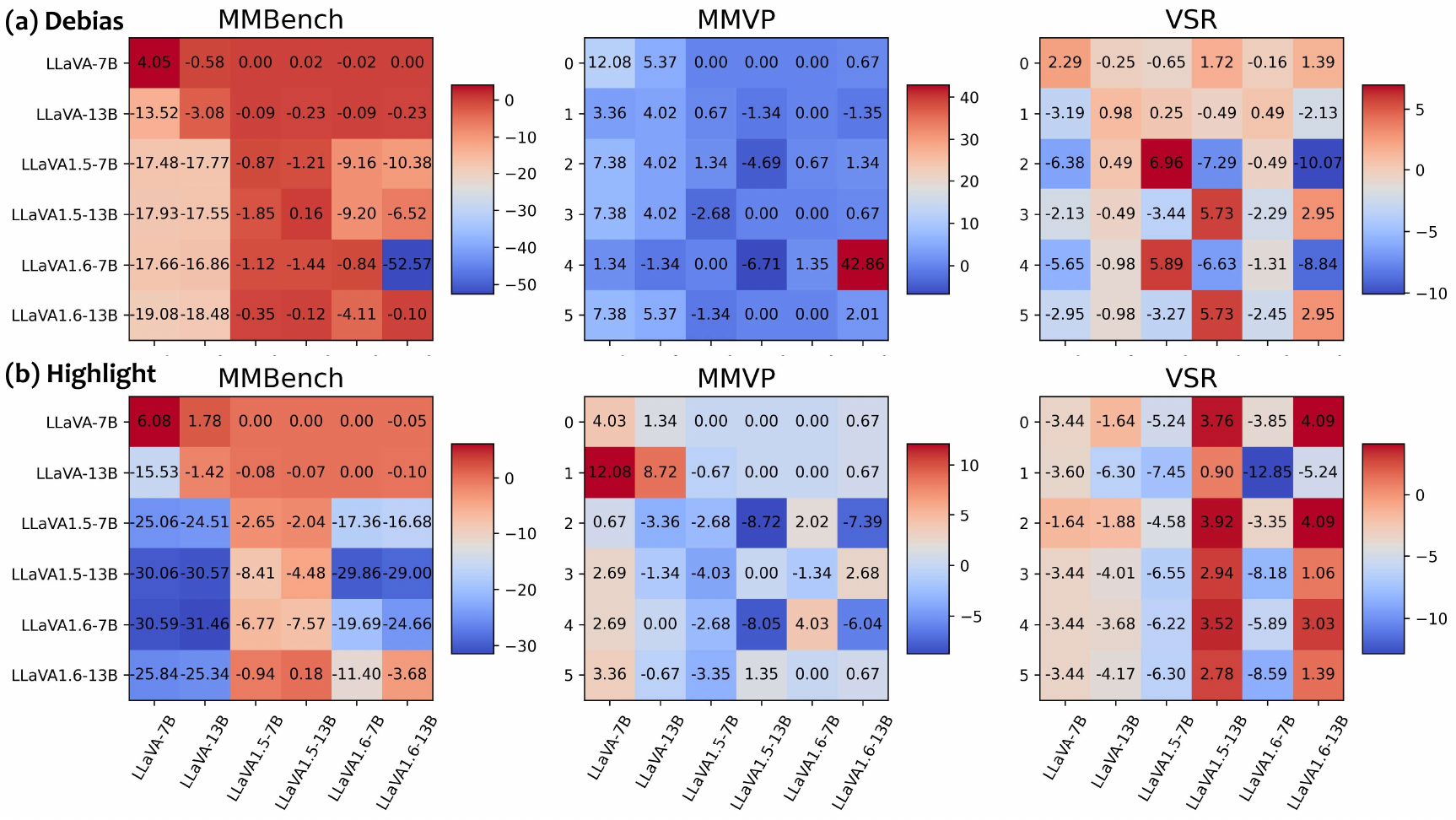}
    \caption{This figure illustrates the results of applying \textit{debias} and \textit{highlight} on model A and model B, which are different models. More details could be found in Sec.\ref{sec:heatmap}. The x-axis represents model B and the y-axis represents model A.}
    \vspace{-1em}
    \label{fig:heatmap}
\end{figure}

\vspace{1em}
\noindent
\textbf{Applying \textit{self-composition} between different models.}
\label{sec:heatmap}
To investigate how \textit{debias} and \textit{highlight} works between different models, \emph{e.g.}, subtract likelihood distribution derived by model A using $\text{Prompt}_{noimg}$ from the likelihood distribution derived by another model B using $\text{Prompt}_{simple}$. In Fig.\ref{fig:heatmap}, the x-axis represents model B and y-axis represents model A, the value is the difference between the performance of applying \textit{debias} or \textit{highlight} on model A and model B and the performance of model B.\footnote{$\alpha$ is set to 1.0 in all experiments in Fig.\ref{fig:heatmap}.} Thus, a higher value in the heatmap means $\textit{debias}$ or $\textit{highlight}$ works well between model A and model B.

As we can see, positive values and small negative values appear on the upper right part of the heatmap, \emph{i.e.}, when model B is relatively better than model A, the method we mentioned above can work. Also, the highest value is always derived from two adjacent models on the coordinate axis, \emph{e.g.}, in subfigure (a), the highest value is derived between LLaVA1.6-7B and LLaVA1.6-13B. In summary, \textbf{when applying \textit{self-composition} method between two different models, if these two models are similar on the downstream task performance, it may work.}

\begin{figure}[thb]
    \centering
    \includegraphics[width=\linewidth]{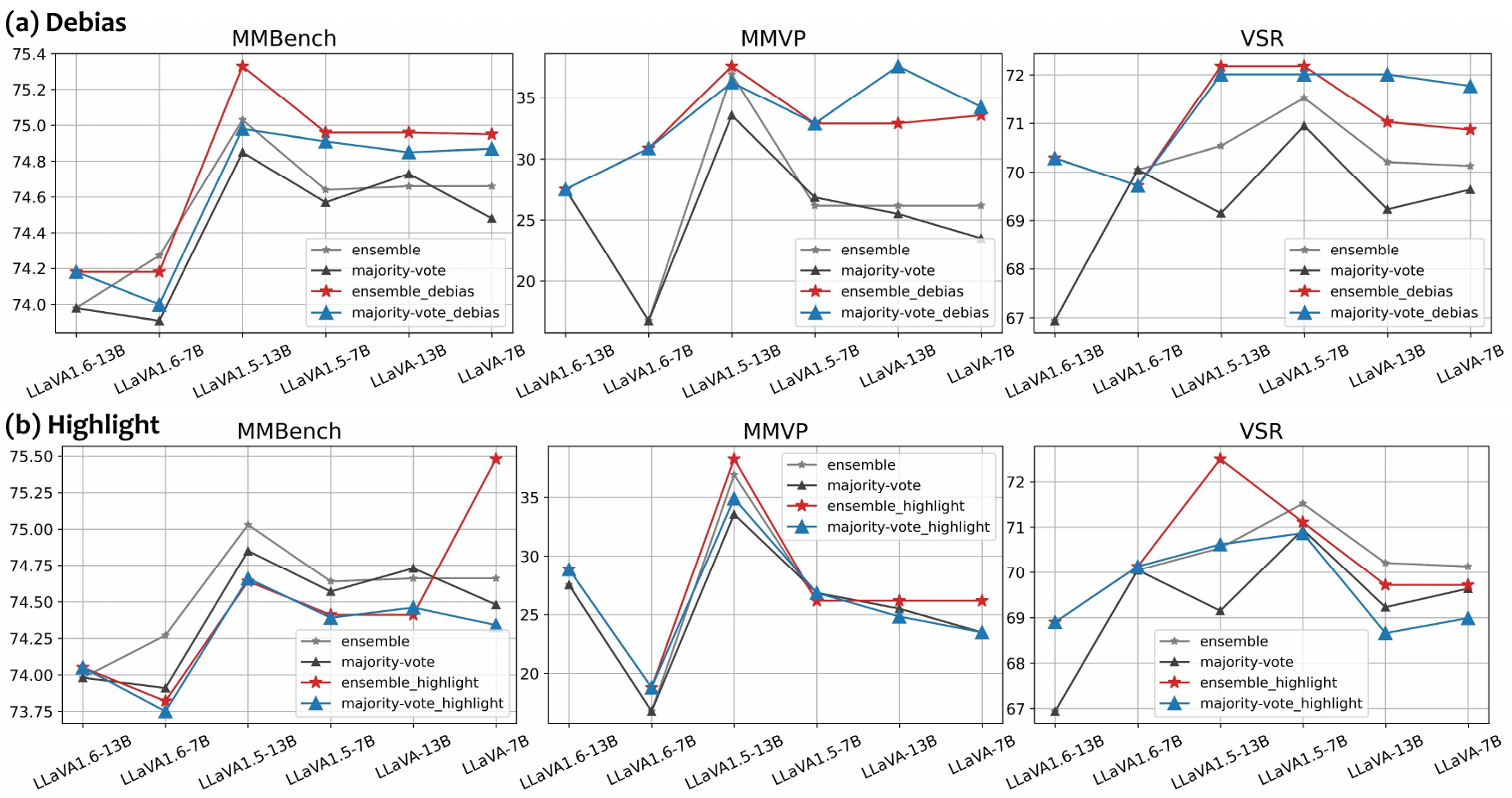}
    \caption{We change the number of models used for \textit{mutual-composition} and \textit{mix-composition}. At each position of the x-axis, we show the results of applying \textit{mutual-composition} and \textit{mix-composition} on the model at this position and the models left to it. It is obvious that models' quality is more important than models' quantity.}
    \label{fig:nums}
\end{figure}

\noindent
\textbf{Adjusting the Number of Models to Ensemble.}
When conducting experiments on LLaVA series, we apply \textit{mutual-composition} and \textit{mix-composition} on all the 6 models. However, fusing all the models may not be the best choice. Thus, we reduce the models to fuse and show the performance change in Fig.\ref{fig:nums}. In this figure, each datapoint represents the result of applying \textit{mutual-composition} or \textit{mix-composition} to the model of the datapoint's x-coordinate and all the models to its left.

As we can see, in most cases, the best performance does not appear at the far right but at ``LLaVA1.5-13B'' \emph{i.e.}, only fusing LLaVA1.5-13B, LLaVA1.6-7B and LLaVA1.6-13B works better than fusing all 6 models. Thus, when applying \textit{mutual-composition} and \textit{mix-composition}, \textbf{models' quality is more important than models' quantity.}

\section{Conclusion}
In this work, we propose ``likelihood composition'', a framework unifying some operation in the model fusing field. Based on this framework, we further propose ``mix-composition'', mixing the ``self-composition'' and ``mutual-composition''. In our experiments, we find ``self-composition'' can boost the MLM significantly on VQA tasks and ``mix-composition'' also bring significant improvement compared with ``mutual-composition''. 

\section{Limitations}
In our work, we did not consider closed source MLMs. For example, we can prompt closed-source MLMs to give their confidence on the list of answers and utilize these likelihood distributions in our proposed composition methods.

\bibliography{custom}

\newpage
\appendix

\section{Appendix}
\label{sec:appendix}

\subsection{Answer with no images.}
\begin{figure}[th]
    \centering
    \includegraphics[width=\linewidth]{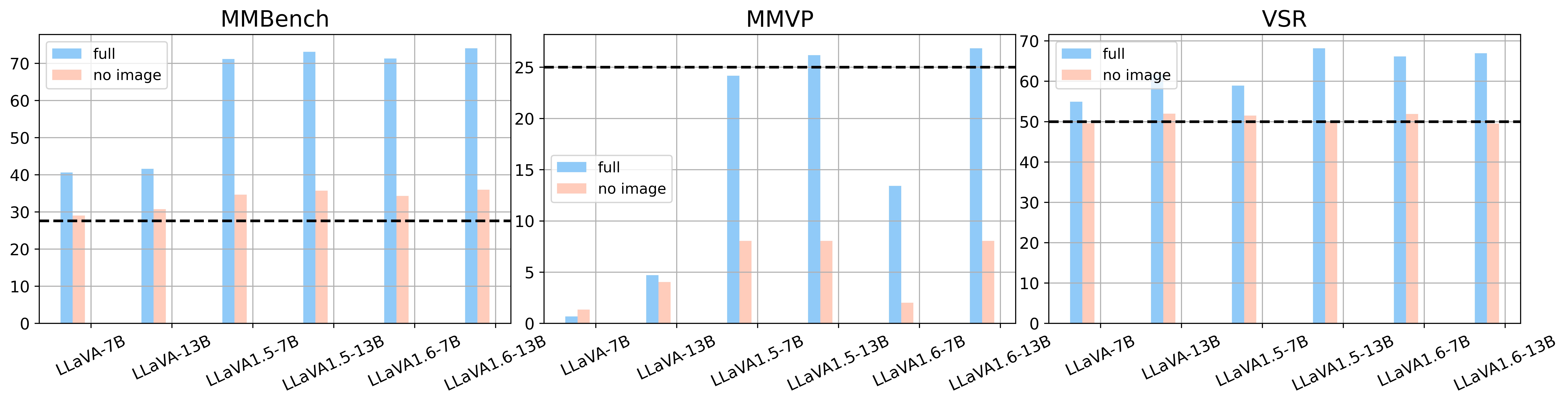}
    \caption{We investigate models' performance when not providing the image. In the figure, the black dashed line represents the random choice score, ``full'' represents the normal case and ``no image'' represents not inputting the image to the model when doing VQA tasks. What's interesting is that models with higher multi-modal understanding ability ``guess'' better when not inputting the image.}
    \label{fig:noimg}
    \vspace{-2em}
\end{figure}

In our proposed \textit{debias} method, we make the model to produce likelihood distribution conditioned only on the question and choices, with no image provided. We can select the predicted answer based this likelihood distribution in the absence of the image. The results is shown in Fig.\ref{fig:noimg}.

We find that the performance of the model with no image provided increases with the increases of the performace conditioned on the image, which is so interesting. \textbf{It seems that with the model's multi-modal understanding ability increasing, the model's ability of ``guessing correctly'' increases also.}

\subsection{Statistics of Datasets}

In Table~\ref{tab:staticdataset}, the statistics of each benchmark, including version and number of samples, are listed.
\begin{table}[h]
    \centering
    \footnotesize
    \adjustbox{max width=\linewidth}{
    \begin{tabular}{l|c|c}
    \toprule
    Benchmark & Version & Number of Samples \\
    \midrule
    MME&-&2373 \\
    MMBench&dev&4377 \\
    MMVP & - &300 \\
    POPE&Popular,Random,Adversarial&8910 \\
    VSR&-&1222\\
    OKVQA&ReForm-Eval\cite{li2023reform}&504 \\
    VQAv2&ReForm-Eval&2144 \\
    Vizwiz&ReForm-Eval&431 \\
    GQA&ReForm-Eval&1257 \\
    \bottomrule
    \end{tabular}}
    \caption{Statistics of 
 each benchmark.}
    \label{tab:staticdataset}
\end{table}

\subsection{Full Results of Debias and Highlight with Different $\alpha$'s Values}

In Table.\ref{tab:llava_series_self}, we show the full results of applying \textit{debias} and \textit{highlight} on LLaVA~\cite{liu2023llava,liu2023improvedllava,liu2024llavanext} series with different $\alpha$'s values. 
\begin{table*}[th]
\vspace{-1em}
    \centering
    \adjustbox{max width=\textwidth}{
    \begin{tabular}{l|c|cc|ccccccccc}
    \toprule
    &$\alpha$&\makecell{De\\bias}&\makecell{High\\light}&MME&MMVP&MMBench&VSR&POPE&VQAv2*&Vizwiz*&GQA*&OKVQA* \\
    \midrule
         \multirow{9}{*}{7B}
         &&&&991.26&0.67&40.62&54.91&61.83&38.99&36.66&36.83&31.94 \\
         \cline{2-13}
         &\multirow{2}{*}{10}
         &\checkmark&&\textbf{1204.57}&	12.08	&44.46	&\textbf{60.47}	&67.85	&41.14	&37.82	&\textbf{39.46}	&31.15  \\
          &&&\checkmark&980.52&	2.68	&43.82&	51.47&	68.90	&36.38&	27.75&	34.37&	28.77 \\
         \cline{2-13}
         &\multirow{2}{*}{1.0}
         &\checkmark&&1069.56&\textbf{12.75}&44.67&57.20&\textbf{76.76}&\textbf{42.07}&38.52&38.19&\textbf{32.54}  \\
         &&&\checkmark&973.94&4.70&\textbf{46.70}&51.47&70.82&38.53&32.02&35.24&31.15  \\
         \cline{2-13}
        &\multirow{2}{*}{0.5}
        &\checkmark&&1036.94	&10.07	&43.71&	56.38	&70.50&	40.67	&\textbf{38.75}	&38.58	&31.94 \\
         &&&\checkmark&971.97&	6.71	&45.72&	52.78&	69.72&	39.23	&35.73	&35.08	&31.35 \\
         \cline{2-13}
         
         &\multirow{2}{*}{0.1}
         &\checkmark&&988.63&2.01&41.81&55.65&61.76&39.69&36.89&37.07&32.34 \\
         &&&\checkmark&987.84&4.03&42.86&59.33&63.35&39.97&36.43&36.20&31.94 \\
         \midrule
         \multirow{9}{*}{13B}
        &&&&1106.00&4.70&41.58&61.62&55.72&37.87&31.79&38.58&28.77 \\
        \cline{2-13}
                 &\multirow{2}{*}{10}
                 &\checkmark&&1190.97	&8.05	&32.81	&62.93&	\textbf{62.13}	&37.22&	30.39	&35.24	&26.79 \\
         &&&\checkmark&\textbf{1248.54}&	14.09	&32.60	&52.13	&53.50	&33.3&	22.04	&33.09&	23.41 \\
         \cline{2-13}
        &\multirow{2}{*}{1.0}
         &\checkmark&&1124.50&8.72&38.50&62.60&59.46&38.20&32.95&37.07&29.17 \\
         &&&\checkmark&1197.31&13.42&40.16&55.32&54.64&34.10&25.06&33.09&25.00  \\
         \cline{2-13}
                  &\multirow{2}{*}{0.5}
                  &\checkmark&&1099.25	&8.72&	41.95&	\textbf{62.68}&	58.59&	\textbf{39.51}&	\textbf{35.50}	&36.75&	\textbf{30.75} \\
        &&&\checkmark&1156.91	&\textbf{14.77}	&42.22&	58.10	&57.28	&34.93&	27.38	&34.37&	25.40 \\
         \cline{2-13}
         &\multirow{2}{*}{0.1}
         &\checkmark&&1090.73&8.72&\textbf{45.97}&61.54&57.00&39.46&34.11&\textbf{38.98}&30.56 \\
         &&&\checkmark&1114.30&8.72&41.88&60.31&58.83&36.99&31.32&36.04&27.78 \\
         \midrule
         \multirow{9}{*}{v1.5-7B}
        &&&&1741.14&24.16&71.17&58.92&85.78&73.09&64.04&65.08&73.81 \\
        \cline{2-13}
                 &\multirow{2}{*}{10}
                 &\checkmark&& 1626.79&	23.49	&66.51&	\textbf{66.69}	&68.56&	\textbf{73.88}&	62.65&	63.33&	72.22\\
         &&&\checkmark&1674.43&	20.81&	61.78&	52.78	&74.33&	67.35&	56.84	&62.29&	61.31 \\
         \cline{2-13}
        &\multirow{2}{*}{1.0}
         &\checkmark&&1723.09&\textbf{25.50}&70.30&65.88&70.19&73.60&63.34&64.52&73.41  \\
         &&&\checkmark&\textbf{1804.74}&21.48&68.52&54.34&73.82&71.83&61.95&64.52&68.85 \\
         \cline{2-13}
                  &\multirow{2}{*}{0.5}
                  &\checkmark&&1736.78&	24.83	&70.98	&63.58	&77.18	&73.69&	63.57&	64.92	&73.61 \\
         &&&\checkmark&1802.38	&22.15&	69.61	&55.97&	73.84	&72.48	&62.41	&65.23	&70.04 \\
         \cline{2-13}
&\multirow{2}{*}{0.1}
         &\checkmark&&1747.96&24.83&\textbf{71.42}&60.80&\textbf{85.95}&73.13&\textbf{64.27}&\textbf{65.31}&\textbf{73.81} \\
         &&&\checkmark&1756.19&22.82&71.05&58.02&73.43&73.13&64.04&65.08&72.22 \\
         \midrule
         \multirow{9}{*}{v1.5-13B}
        &&&&1782.34&26.17&73.09&68.17&84.70&75.70&75.17&67.70&76.19 \\
        \cline{2-13}
                 &\multirow{2}{*}{10}
                 &\checkmark&&1796.79&	26.85	&72.17	&73.08	&71.06&	75.33&	73.78	&67.14&	74.60 \\
        &&&\checkmark& 1740.95&	\textbf{28.19}	&55.36	&61.54&	58.88	&70.85&	73.32	&65.31&	60.71\\
         \cline{2-13}
        &\multirow{2}{*}{1.0}
         &\checkmark&&\textbf{1833.70}&26.17&73.25&\textbf{73.90}&79.83&75.42&\textbf{75.64}&67.70&75.40  \\
         &&&\checkmark&1819.68&26.17&68.61&71.11&59.11&75.79&75.17&66.27&72.22  \\
         \cline{2-13}
                  &\multirow{2}{*}{0.5}
                  &\checkmark&& 1805.9&	26.17	&\textbf{73.27}	&72.83	&84.55	&75.51	&75.41	&67.86	&75.60\\
         &&&\checkmark&1791.54	&25.50&	71.21&	72.42	&59.27	&76.31	&75.64	&67.14	&73.21 \\
         \cline{2-13}
         &\multirow{2}{*}{0.1}
         &\checkmark&&1789.38&26.85&73.22&69.89&\textbf{86.04}&75.75&75.41&67.70&75.79 \\
         &&&\checkmark&1779.75&25.50&72.97&70.29&60.56&\textbf{75.84}&75.14&\textbf{68.10}&\textbf{76.19} \\
         \midrule
         \multirow{9}{*}{v1.6-7B}
        &&&&1691.81&13.42&\textbf{71.30}&66.12&67.33&67.07&54.52&59.35&\textbf{72.62} \\
        \cline{2-13}
                 &\multirow{2}{*}{10}
                 &\checkmark&& 1653.97	&16.78	&68.59&	58.35	&72.38	&66.09	&54.76	&57.84	&70.83\\
         &&&\checkmark&1679.73	&17.45&	34.73	&51.55	&61.17	&50.00	&29.00	&52.11&	58.33 \\
         \cline{2-13}
        &\multirow{2}{*}{1.0}
         &\checkmark&&\textbf{1765.97}&14.77&70.46&64.81&71.11&66.84&54.76&58.31&71.43  \\
         &&&\checkmark&1679.07&\textbf{17.45}&51.61&60.23&60.61&67.07&52.90&\textbf{59.51}&68.85  \\
         \cline{2-13}
                  &\multirow{2}{*}{0.5}
         &\checkmark&& 1754.84	&14.77	&70.92	&\textbf{66.53}	&70.54&	67.02&	54.99&	58.55&	72.22\\
         &&&\checkmark&1692.89&	16.11	&59.33&	62.44&	60.57	&67.07&52.90&59.51&			71.23  \\
         \cline{2-13}
         &\multirow{2}{*}{0.1}
         &\checkmark&&1711.07&13.42&71.14&66.45&68.50&67.16&54.99&59.11&72.42 \\
         &&&\checkmark&1703.37&14.09&70.41&65.96&\textbf{72.42}&\textbf{69.26}&\textbf{67.02}&59.19&72.42\\
         \midrule
         \multirow{9}{*}{v1.6-13B}
        &&&&1807.45&26.85&74.05&66.94&84.74&76.21&\textbf{81.67}&66.75&75.99 \\
        \cline{2-13}
                 &\multirow{2}{*}{10}
                 &\checkmark&& 1803.29	&\textbf{32.21}	&71.30	&68.09&	76.65	&73.65	&71.00&	67.14&73.02	\\
         &&&\checkmark&1723.66&	27.52&	61.96&	53.60&	56.70	&71.88	&73.55	&\textbf{68.26}	&69.64 \\
         \cline{2-13}
        &\multirow{2}{*}{1.0}
         &\checkmark&&1790.79&28.86&73.95&69.89&81.36&75.70&78.42&67.46&75.60  \\
         &&&\checkmark&1726.69&27.52&70.37&68.33&56.83&75.89&77.73&68.18&74.60  \\
         \cline{2-13}
                  &\multirow{2}{*}{0.5}
                  &\checkmark&& 1787.40&	25.50&	\textbf{74.18}&	70.29&	83.79	&76.07&	80.04	&67.46&76.59	\\
         &&&\checkmark& 1746.43&	28.19	&72.38	&\textbf{70.79}	&57.03&	\textbf{76.49}&	79.81	&67.70&	74.80\\
         \cline{2-13}
         &\multirow{2}{*}{0.1}
         &\checkmark&&1800.68&28.19&74.11&67.92&\textbf{84.78}&76.35&80.97&67.06&\textbf{76.59} \\
         &&&\checkmark&\textbf{1814.42}&28.86&74.05&68.90&59.47&76.21&81.21&66.91&76.19 \\
    \bottomrule
    \end{tabular}}
    \caption{Full results of \textit{debias} and \textit{highlight}.}
    \label{tab:llava_series_self}
    \vspace{-3em}
\end{table*}

\end{document}